\documentclass{article}

\usepackage{arxiv}

\usepackage[utf8]{inputenc} % allow utf-8 input
\usepackage[T1]{fontenc}    % use 8-bit T1 fonts
\usepackage{hyperref}       % hyperlinks
\usepackage{url}            % simple URL typesetting
\usepackage{booktabs}       % professional-quality tables
\usepackage{amsfonts}       % blackboard math symbols
\usepackage{nicefrac}       % compact symbols for 1/2, etc.
\usepackage{microtype}      % microtypography
\usepackage{lipsum}

\usepackage{graphicx}
\usepackage{amsmath,amssymb} % define this before the line numbering.
\usepackage{color}
\usepackage{subcaption}
\usepackage{bm}

\makeatletter
\newcommand{\thickhline}{%
    \noalign {\ifnum 0=`}\fi \hrule height 1pt
    \futurelet \reserved@a \@xhline
}

\title{A New Look at Ghost Normalization}

\author{
  Neofytos Dimitriou\\
  School of Computer Science\\
  University of St Andrews\\
  St Andrews, United Kingdom \\
  \texttt{neofytosd@gmail.com} \\
   \And
  Ognjen Arandjelovi\'c\ \\
  School of Computer Science\\
  University of St Andrews\\
  St Andrews, United Kingdom \\
  \texttt{ognjen.arandjelovic@gmail.com} \\
}

\begin{document}
\maketitle

\begin{abstract}
Batch normalization (BatchNorm) is an effective yet poorly understood technique for neural network optimization. It is often assumed that the degradation in BatchNorm performance to smaller batch sizes stems from it having to estimate layer statistics using smaller sample sizes. However, recently, Ghost normalization (GhostNorm), a variant of BatchNorm that explicitly uses smaller sample sizes for normalization, has been shown to improve upon BatchNorm in some datasets. Our contributions are: (i) we uncover a source of regularization that is unique to GhostNorm, and not simply an extension from BatchNorm, (ii) three types of GhostNorm implementations are described, two of which employ BatchNorm as the underlying normalization technique, (iii) by visualising the loss landscape of GhostNorm, we observe that GhostNorm consistently decreases the smoothness when compared to BatchNorm, (iv) we introduce Sequential Normalization (SeqNorm), and report superior performance over state-of-the-art methodologies on both CIFAR--10 and CIFAR--100 datasets.
\end{abstract}

% keywords can be removed
\tiny
\normalsize

\section{Introduction}
The effectiveness of Batch Normalization (BatchNorm), a technique first introduced by Ioeffe and Szegedy~\cite{IoffSzeg2015} on neural network optimization has been demonstrated over the years on a variety of tasks, including computer vision,~\cite{Krizhevsky2017,HuanLiuWein2016,HeZhanRenSun2016}, speech recognition~\cite{Graves2013}, and other~\cite{Ilya2014,Silver2016,Mnih2015}. BatchNorm is typically embedded at each neural network (NN) layer either before or after the activation function, normalizing and projecting the input features to match a Gaussian-like distribution. Consequently, the activation values of each layer maintain more stable distributions during NN training which in turn is thought to enable faster convergence and better generalization performances~\cite{IoffSzeg2015,SantTsipIlyaMadr2018,Nils2018}. 

Despite the wide adoption and practical success of BatchNorm, its underlying mechanics within the context of NN optimization has yet to be fully understood. Initially, Ioeffe and Szegedy suggested that it came from it reducing the so-called internal covariate shift~\cite{IoffSzeg2015}. At a high level, internal covariate shift refers to the change in the distribution of the inputs of each NN layer that is caused by updates to the previous layers. This continual change throughout training was conjectured to negatively affect optimization~\cite{IoffSzeg2015,SantTsipIlyaMadr2018}. However, recent research disputes that with compelling evidence that demonstrates how BatchNorm may in fact be increasing internal covariate shift~\cite{SantTsipIlyaMadr2018}. Instead, the effectiveness of BatchNorm is argued to be a consequence of a smoother loss landscape~\cite{SantTsipIlyaMadr2018}.

Following the effectiveness of BatchNorm on NN optimization, a number of different normalization techniques have been introduced~\cite{Sergey2017,LeiB2016,Qiao2019,WuHe2018,UlyaVedaLemp2016}. Their main inspiration was to provide different ways of normalizing the activations without being inherently affected by the batch size. In particular, it is often observed that BatchNorm performs worse with smaller batch sizes~\cite{Sergey2017,WuHe2018,YanWanZhanZhan+2020}. This degradation has been widely associated to BatchNorm computing poorer estimates of mean and variance due to having a smaller sample size. However, recent demonstration of the effectiveness of GhostNorm comes in antithesis with the above belief~\cite{SummDinn2020}. GhostNorm explicitly divides the mini--batch into smaller batches and normalizes over them independently~\cite{Elad2017}. Nevertheless, when compared to other normalization techniques~\cite{Sergey2017,LeiB2016,Qiao2019,WuHe2018,UlyaVedaLemp2016}, the adoption of GhostNorm has been rather scarce, and narrow to large batch size training regimes~\cite{SummDinn2020}.

In this work, we take a new look at GhostNorm, and contribute in the following ways: (i) Identifying a source of regularization that is unique to GhostNorm, and discussing the difference against other normalization techniques, (ii) Providing a direct way of implementing GhostNorm, as well as through the use of accumulating gradients and multiple GPUs, (iii) Visualizing the loss landscape of GhostNorm under vastly different experimental setups, and observing that GhostNorm consistently decreases the smoothness of the loss landscape, especially on the later epochs of training, (iv) Introducing SeqNorm as a new normalization technique, (v) Surpassing the performance of baselines that are based on state-of-the-art (SOTA) methodologies on CIFAR--10 and CIFAR--100 for both GhostNorm and SeqNorm, with the latter even surpassing the current SOTA on CIFAR--100 that employs a data augmentation strategy.

\subsection{Related Work}
Ghost Normalization is a technique originally introduced by Hofer \textit{et al}.\ \cite{Elad2017}. Over the years, the primary use of GhostNorm has been to optimize NNs with large batch sizes and multiple GPUs~\cite{SummDinn2020}. However, when compared to other normalization techniques~\cite{Sergey2017,LeiB2016,Qiao2019,WuHe2018,UlyaVedaLemp2016}, the adoption of GhostNorm has been rather scarce.

In parallel to our work, we were able to identify a recent published work that has in fact experimented with GhostNorm on both small and medium batch size training regimes. Summers and Dinnen~\cite{SummDinn2020} tuned the number of groups within GhostNorm (see section~\ref{sec:formulation}) on CIFAR--100, Caltech--256, and SVHN, and reported positive results on the first two datasets. More results are reported on other datasets through transfer learning, however, the use of other new optimization methods confound the contribution made by GhostNorm.

The closest line of work to SeqNorm is, again, found in the work of Summers and Dinnen~\cite{SummDinn2020}. Therein they employ a normalization technique which although at first glance may appear similar to SeqNorm, at a fundamental level, it is rather different. This stems from the vastly different goals between our works, i.e.\ they try to increase the available information when small batch sizes are used~\cite{SummDinn2020}, whereas we strive to improve GhostNorm in the more general setting. At a high level, where SeqNorm performs GroupNorm and GhostNorm sequentially, their normalization method applies both simultaneously. At a fundamental level, the latter embeds the stochastic nature of GhostNorm (see section~\ref{sec:effects_gn}) into that of GroupNorm, thereby potentially disrupting the learning of channel grouping within NNs. Switchable normalization is also of some relevance to SeqNorm as it enables the NN to learn which normalization techniques to employ at different layers~\cite{LuoRenPengZhang+2019}. However, similar to the previous work, simultaneously applying different normalization techniques has a fundamentally different effect than SeqNorm.

Related to our work is also research geared towards exploring the effects of BatchNorm on optimization~\cite{Nils2018,SantTsipIlyaMadr2018,Jonas2019}. Finally, of some relevance is also the large body of work that exists on improving BatchNorm at the small batch training regime~\cite{Sergey2017,WuHe2018,LuoZhanWangGao2020,YanWanZhanZhan+2020}.

\section{Methodology}
\label{sec:methods}

\subsection{Formulation}
\label{sec:formulation}

\begin{figure}
\centering
\includegraphics[height=72mm]{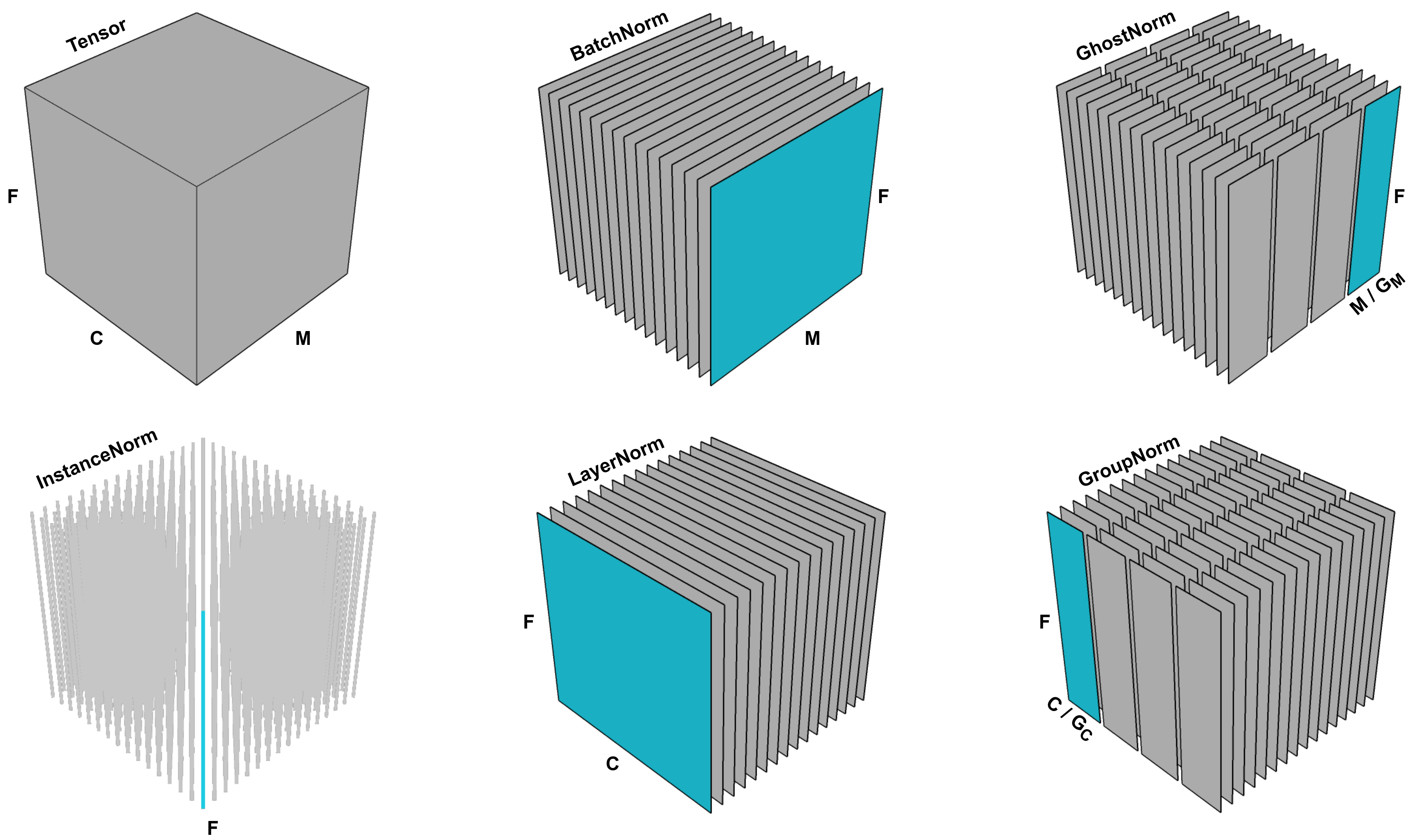}
\caption{
The input tensor is divided into a number of line (1D) or plane (2D) slices. Each normalization technique slices the input tensor differently and each slice is normalized independently of the other slices.
}
\label{fig:norm_technques}
\end{figure}

\sloppy
Given a fully-connected or convolutional neural network, the parameters of a typical layer $l$ with normalization, $Norm$, are the weights $W^l$ as well as the scale and shift parameters $\gamma^{l}$ and $\beta^{l}$. For brevity, we omit the $l$ superscript. Given an input tensor $X$, the activation values $A$ of layer $l$ are computed as:
% \in\mathbb{R}^{m \times F}
\begin{align}
  A = g (Norm(X \odot W) \otimes \gamma + \beta)
\label{eq:1}
\end{align}
where $g(\cdot)$ is the activation function, $\odot$ corresponds to either matrix multiplication or convolution for fully-connected and convolutional layers respectively, and $\otimes$ describes an element-wise multiplication. 

Most normalization techniques differ in how they normalize the product ${X \odot W}$. Let the product be a tensor with ${(M, C, F)}$ dimensions where $M$ is the so-called mini--batch size, or just batch size, $C$ is the channels dimension, and $F$ is the spatial dimension. 
% Although the normalization techniques we discuss below can employ tensors of different dimensions, the actual normalization still takes place on $M$ and $C$, i.e.\ typically the first two dimensions.
% Nevertheless, normalization still takes place on either $M$ or $C$ dimensions.
% even though an input tensor can have a different number of dimensions, the normalization techniques we discuss below, for the most part, either collapse the extra dimensions into $F$ or introduce the missing one. 
% For instance, a 2D image with ($M$, $C$, $H$, $W$) dimensions turns to a tensor with ($M$, $C$, $H \times W$) dimensions.
% BatchNorm normalizes the values of a given tensor (i.e. the product of $X$ and $W$) for each channel dimension independently. 

In \textit{BatchNorm}, the given tensor is normalized across the channels dimension. In particular, the mean and variance are computed across $C$ number of slices of ${(M, F)}$ dimensions (see Figure~\ref{fig:norm_technques}) which are subsequently used to normalize each channel $c\in C$ independently. In \textit{LayerNorm}, statistics are computed over $M$ slices of ${(C, F)}$ dimensions, normalizing the values of each data sample $m\in M$ independently. \textit{InstanceNorm} normalizes the values of the tensor over both $M$ and $C$, i.e.\ computes statistics across ${M \times C}$ slices of $F$ dimensions.

\textit{GroupNorm} can be thought as an extension to LayerNorm wherein the $C$ dimension is divided into $G_C$ number of groups, i.e.\ ${(M, G_C, ^C/_{G_C}, F)}$. Statistics are calculated over $M \times G_C$ slices of ${(^C/_{G_C}, F)}$ dimensions. Similarly, \textit{GhostNorm} can be thought as an extension to BatchNorm, wherein the $M$ dimension is divided into $G_M$ groups, normalizing over $C \times G_M$ slices of ${(^M/_{G_M}, F)}$ dimensions. Both $G_C$ and $G_M$ are hyperparameters that can be tuned based on a validation set. All of the aforementioned normalization techniques are illustrated in Figure~\ref{fig:norm_technques}.

\textit{SeqNorm} employs both GroupNorm and GhostNorm in a sequential manner. Initially, the input tensor is divided into ${(M, G_C, ^C/_{G_C}, F)}$ dimensions, normalizing across $M \times G_C$ number of slices, i.e.\ same as GroupNorm. Then, once the $G_C$ and $^C/_{G_C}$ dimensions are collapsed back together, the input tensor is divided into ${(G_M, ^M/_{G_M}, C, F)}$ dimensions for normalizing over ${C \times G_M}$ slices of ${(^M/_{G_M}, F)}$ dimensions. 
% In essence, two normalization techniques are applied instead of one.

Any of the slices described above is treated as a set of values $S$ with one dimension. The mean and variance of $S$ are computed in the traditional way (see Equation~\ref{eq:2}). The values of $S$ are then normalized as shown below.
\begin{align}
\label{eq:2}
  \mu & = \frac{1}{M} \sum_{x \in S} x \text{ and } \sigma^2 = \frac{1}{M} \sum_{x \in S} (x - \mu)^2
\end{align}

\begin{align}
\label{eq:3}
\forall x \in S \text{, } x = \frac{x - \mu}{\sqrt{\sigma^2 + \epsilon}}
\end{align}

Once all slices are normalized, the output of the $Norm$ layer is simply the concatenation of all slices back into the initial tensor shape.
\fussy
%------------------------------------------------------------------------- 
\subsection{The effects of Ghost Normalization}
\label{sec:effects_gn}
There is only one other published work which has investigated the effectiveness of Ghost Normalization for small and medium mini-batch sizes~\cite{SummDinn2020}. Therein, they hypothesize that GhostNorm offers stronger regularization than BatchNorm as it computes the normalization statistics on smaller sample sizes~\cite{SummDinn2020}. In this section, we support that hypothesis by providing insights into a particular source of regularization, unique to GhostNorm, that stems from normalizing groups of activations during a forward pass.  

\sloppy
Consider as an example the tuple $X$ with ${(35, 39, 30, 4, 38, 26, 27, 19)}$ values which can be thought as an input tensor with ${(8, 1, 1)}$ dimensions. Given to a BatchNorm layer, the output is the normalized version $\bar{X}$ with values  ${(0.7,1.1,0.3,-2.2, 1.0,-0.1,-0.02,-0.8)}$. Note how although the values have changed, the ranking order of the activation values has remained the same, e.g.\ the 2nd value is larger than the 5th value in both $X$ (${39 > 38}$) and $\bar{X}$ (${1.1 > 1.0}$). More formally, the following holds true:
\begin{align}
\label{eq:4}
\begin{split}
\text{Given n-tuples } X = (x_0, x_1, .. x_n) \text{ and } \bar X = (\bar x_0, \bar x_1, ..., \bar x_n) \text{,}\\
\forall i, j \in I, \bar x_i > \bar x_j \iff x_i > x_j\\
\bar x_i < \bar x_j \iff x_i < x_j\\
\bar x_i = \bar x_j \iff x_i = x_j
\end{split}
\end{align}

On the other hand, given $X$ to a GhostNorm layer with $G_M=2$, the output $\bar X$ is $(0.6,  0.9,  0.2, -1.7, 1.5, -0.2, -0.07, -1.2)$. Now, we observe that after normalization, the 2nd value has become much smaller than the 5th value in $\bar X$ (${0.9 < 1.5}$). Where BatchNorm preserves the ranking order, GhostNorm can modify the importance of each sample, and hence alter the course of optimization. Our experimental results demonstrate how GhostNorm improves upon BatchNorm, supporting the hypothesis that the above type of regularization can be beneficial to optimization. Note that for BatchNorm the condition in Equation~\ref{eq:4} only holds true across the ${M\times F}$ dimension of the input tensor whereas for GhostNorm it cannot be guaranteed for any given dimension.

\paragraph{GhostNorm to BatchNorm}
One can argue that the same type of regularization can be found in BatchNorm over different mini--batches, e.g. given $[35, 39, 30, 4]$ and $[38, 26, 27, 19]$ as two different mini--batches. However, GhostNorm introduces the above during each forward pass rather than between forward passes. Hence, it is a regularization that is embedded during learning (GhostNorm), rather than across learning (BatchNorm).

\fussy

\paragraph{GhostNorm to GroupNorm} Despite the visual symmetry between GhostNorm and GroupNorm, there is one major difference. Grouping has been employed extensively in classical feature engineering, such as SIFT, HOG, and GIST, wherein independent normalization is often performed over these groups~\cite{WuHe2018}. At a high level, GroupNorm can be thought as motivating the network to group similar features together~\cite{WuHe2018}. On the other hand, that learning behaviour is not possible with GhostNorm due to random sampling, and random arrangement of data within each mini--batch. Therefore, we hypothesize that the effects of these two normalization techniques could be combined for their benefits to be accumulated. Specifically, we propose SeqNorm, a normalization technique that employs both GroupNorm and GhostNorm in a sequential manner.

\subsection{Implementation} 
% SeqNorm employs both GroupNorm and GhostNorm. 
\paragraph{Ghost Normalization} 
% Herein, we identify three different ways of implementing GhostNorm (one explicit and two implicit).
The direct approach of implementing GhostNorm is shown in Figure~\ref{fig:code1}. Although the expontential moving averages are ommited for brevity~\footnote
{
The full implementation will be provided in a code repository. 
}, it's worth mentioning that they were accumulated in the same way as BatchNorm. In addition to the above direct implementation, GhostNorm can be effectively employed while using BatchNorm as the underlying normalization technique.
% However, it is important to discuss two other implementation approaches of GhostNorm that in fact employ BatchNorm as the underlying normalization technique.

\begin{figure}
\begin{verbatim}
def GhostNorm(X, groupsM, eps=1e-05):
    """
    X: Input Tensor with (M, C, F) dimensions
    groupsM: Number of groups for the mini-batch dimension
    eps: A small value to prevent division by zero
    """
    # Split the mini-batch dimension into groups of smaller batches
    M, C, F = X.shape
    X = X.reshape(groupsM, -1, C, F)
    
    # Calculate statistics over dim(0) x dim(2) number
    # of slices of dim(1) x dim(3) dimension each
    mean = X.mean([1, 3], keepdim=True)
    var = X.var([1, 3], unbiased=False, keepdim=True)
    
    # Normalize X
    X = (X - mean) / (torch.sqrt(var + self.eps))
    
    # Reshape into the initial tensor shape
    X = X.reshape(M, C, F)
    
    return X
\end{verbatim}
\caption{Python code for GhostNorm in PyTorch.}
\label{fig:code1}
\end{figure}

When the desired batch size exceeds the memory capacity of the available GPUs, practitioners often resort to the use of accumulating gradients. That is, instead of having a single forward pass with $M$ examples through the network, $n_{fp}$ number of forward passes are made with $^M/_{n_{fp}}$ examples each. Most of the time, gradients computed using a smaller number of training examples, i.e.\ $^M/_{n_{fp}}$, and accumulated over a number of forward passes $n_{fp}$ are identical to those computed using a single forward pass of $M$ training examples. However, it turns out that when BatchNorm is employed in the neural network, the gradients can be substantially different for the above two cases. This is a consequence of the mean and variance calculation (see Equation \ref{eq:2}) since each forwarded smaller batch of $^M/_{n_{fp}}$ data will have a different mean and variance than if all $M$ examples were present. Accumulating gradients with BatchNorm can thus be thought as an alternative way of using GhostNorm with the number of forward passes $n_{fp}$ corresponding to the number of groups $G_M$. A PyTorch implementation of accumulating gradients is shown in Figure~\ref{fig:code2}.
% Momentum difference + more forward passes -> slower

\begin{figure}
\begin{verbatim}
def train__for_an_epoch():
    model.train()
    model.zero_grad()
    for i, (X, y) in enumerate(train_loader):
        outputs = model(X)
        loss = loss_function(outputs, y)
        loss = loss / acc_steps
        loss.backward()
        if (i + 1) % acc_steps == 0:
            optimizer.step()
            model.zero_grad()
\end{verbatim}
\caption{Python code for accumulating gradients in PyTorch.}
\label{fig:code2}
\end{figure}

Finally, the most popular implementation of GhostNorm via BatchNorm, albeit typically unintentional, comes as a consequence of using multiple GPUs. Given $n_g$ GPUs and $M$ training examples, $M/n_g$ examples are forwarded to each GPU. If the BatchNorm statistics are not synchronized across the GPUs (i.e.\ Synchnronized BatchNorm ref), which is often the case for image classification, then $n_g$ corresponds to the number of groups $G_M$.

A practitioner who would like to use GhostNorm should employ the implementation shown in Figure~\ref{fig:code1}. Nevertheless, under the discussed circumstances, one could explore GhostNorm through the use of the other implementations. 

% It is worth mentioning that when accumulating gradients or multiple GPUs are employed, the momentum of the moving averages should be calibrated. This comes as a consequence of forwarding tensors sequentially, rather than in parallel, which in turn makes the statistics of each mini--batch more important in the context of moving averages.

\paragraph{Sequential Normalization} The implementation of SeqNorm is straightforward since it combines GroupNorm, a widely implemented normalization technique, and GhostNorm for which we have discussed three possible implementations.

\section{Experiments}
% Additional detail and results are provided for each of the experiments in the Appendix.
In this section, we first strive to take a closer look into GhostNorm by visualizing the smoothness of the loss landscape during training; a component which has been described as the primary reason behind the effectiveness of BatchNorm. Then, we evaluate the effectiveness of both GhostNorm and SeqNorm on the standard image classification datasets of CIFAR--10 and CIFAR--100. Note that in all of our experiments, the smallest $^M/_{G_M}$ we employ for both SeqNorm and GhostNorm is $4$. A ratio of $1$ would be undefined for normalization, whereas a ratio of $2$ results in large information corruption, i.e.\ all activation values are reduced to either $1$~or~$-1$. 

% Also, the maximum $G_C$ is by architectural design set to $16$, i.e.\ the smallest layer in the network has $16$ channels.

\subsection{Loss landscape visualization}
\paragraph{Implementation details}
% We begin our analysis with experiments on the MNIST and CIFAR--10 data set. 
On MNIST, we train a fully-connected neural network (SimpleNet) with two fully-connected layers of $512$ and $300$ neurons. The input images are transformed to one-dimensional vectors of $784$ channels, and are normalized based on the mean and variance of the training set. The learning rate is set to $0.4$ for a batch size of $512$ on a single GPU.
% , and following the common practice~\cite{WuHe2018}, we adjust the learning rate proportionally for smaller batch sizes (specifically, the learning rate is multiplied by $batch\_size/512$). 
% TODO APPENDIX?
% We employ a vanilla optimizer, stochastic gradient descent, with no momentum and no weight decay, and train SimpleNet for $25$ epochs. 
In addition to training SimpleNet with BatchNorm and GhostNorm, we also train a SimpleNet baseline without any normalization technique.

% by training for only $50$ epochs

A residual convolutional network with $56$ layers (ResNet--56)~\cite{HeZhanRenSun2016} is employed for CIFAR--10. We achieve super--convergence by using the one cycle learning policy described in the work of Smith and Topin~\cite{SmitTopi2017}. Horizontal flipping, and pad-and-crop transformations are used for data augmentation. We use a triangularly annealing learning rate schedule between $0.1$ and $3$, and down to $0.0001$ for the last few epochs. In order to train ResNet--56 without a normalization technique (baseline), we had to adjust the cyclical learning rate schedule to ($0.1$, $1$).

For both datasets, we train the networks on $50,000$ training images, and evaluate on $10,000$ testing images. 
% Our hyperparameter values were adopted from relevant previous work or simply set to default values [refs for MNIST and CIFAR10]. 
For completeness, further implementation details are provided in the Appendix. 

\paragraph{Loss landscape} We visualize the loss landscape during optimization using an approach that was described by Santurkar et al.~\cite{SantTsipIlyaMadr2018}. Each time the network parameters are to be updated, we walk towards the gradient direction and compute the loss at multiple points. This enable us to visualise the smoothness of the loss landscape by observing how predictive the computed gradients are. In particular, at each step of updating the network parameters, we compute the loss at a range of learning rates, and store both the minimum and maximum loss. For MNIST, we compute the loss at $8$ learning rates $\in [0.1, 0.2, 0.3, ..., 0.8$], whereas for CIFAR--10, we do so for $4$ cyclical learning rates $\in [(0.05, 1.5), (0.1, 3), (0.15, 4.5), (0.2, 6)]$, and analogously for the baseline.

\begin{figure}
\centering
\begin{subfigure}{.5\textwidth}
  \centering
  \includegraphics[width=0.8\linewidth]{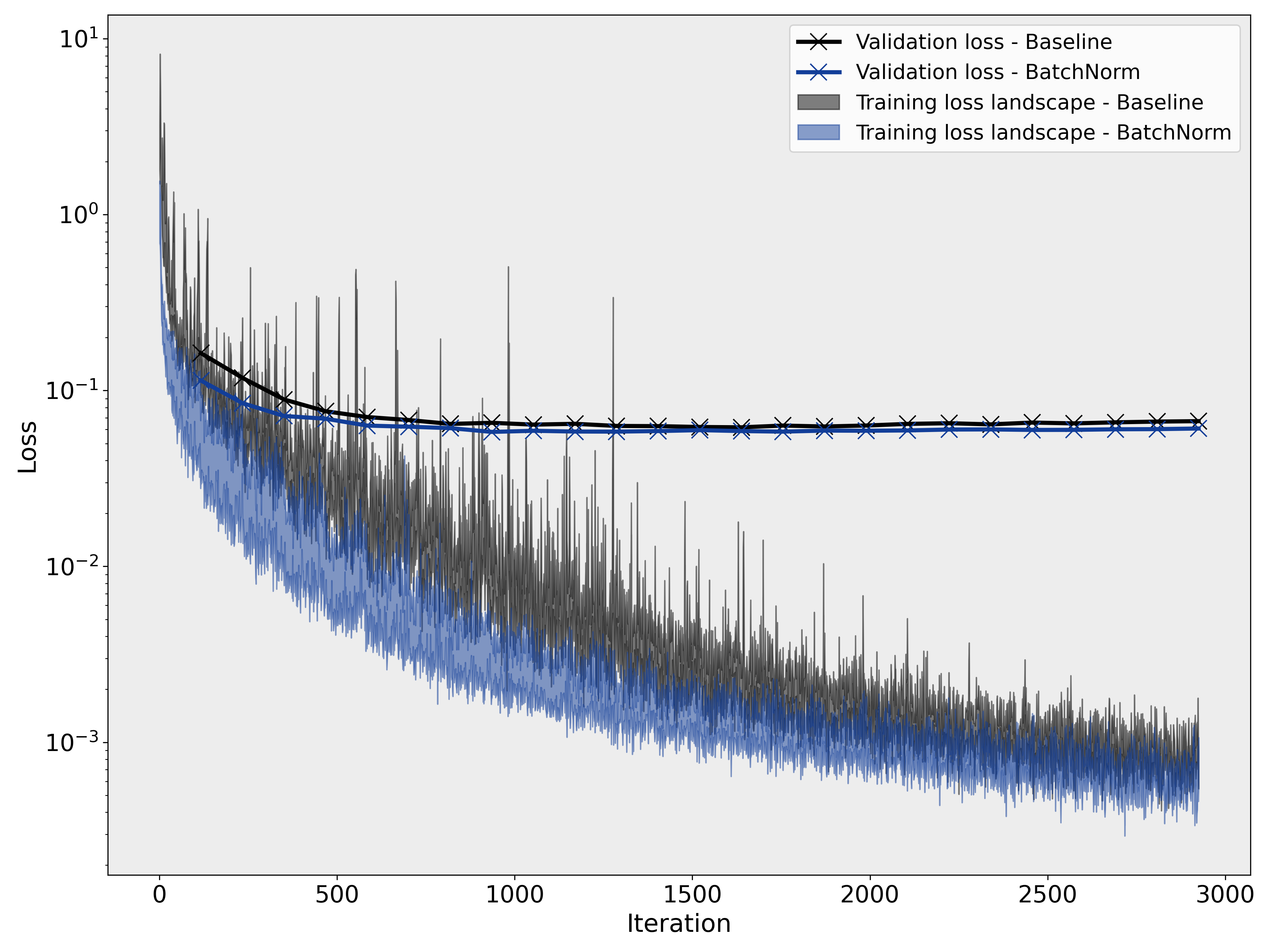}
\end{subfigure}%
\begin{subfigure}{.5\textwidth}
  \centering
  \includegraphics[width=0.8\linewidth]{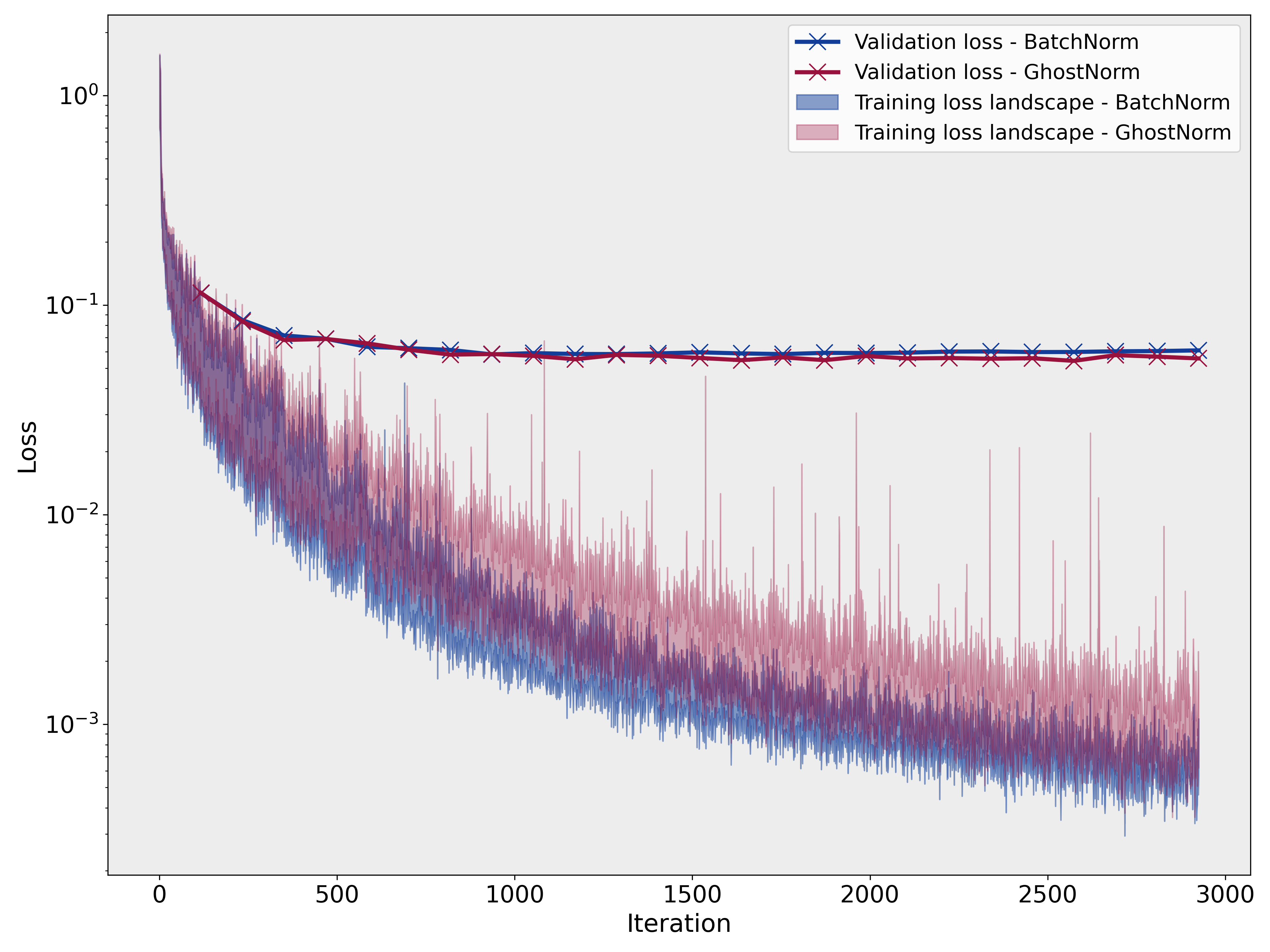}
\end{subfigure}
\vskip\baselineskip
\begin{subfigure}{.5\textwidth}
  \centering
  \includegraphics[width=0.8\linewidth]{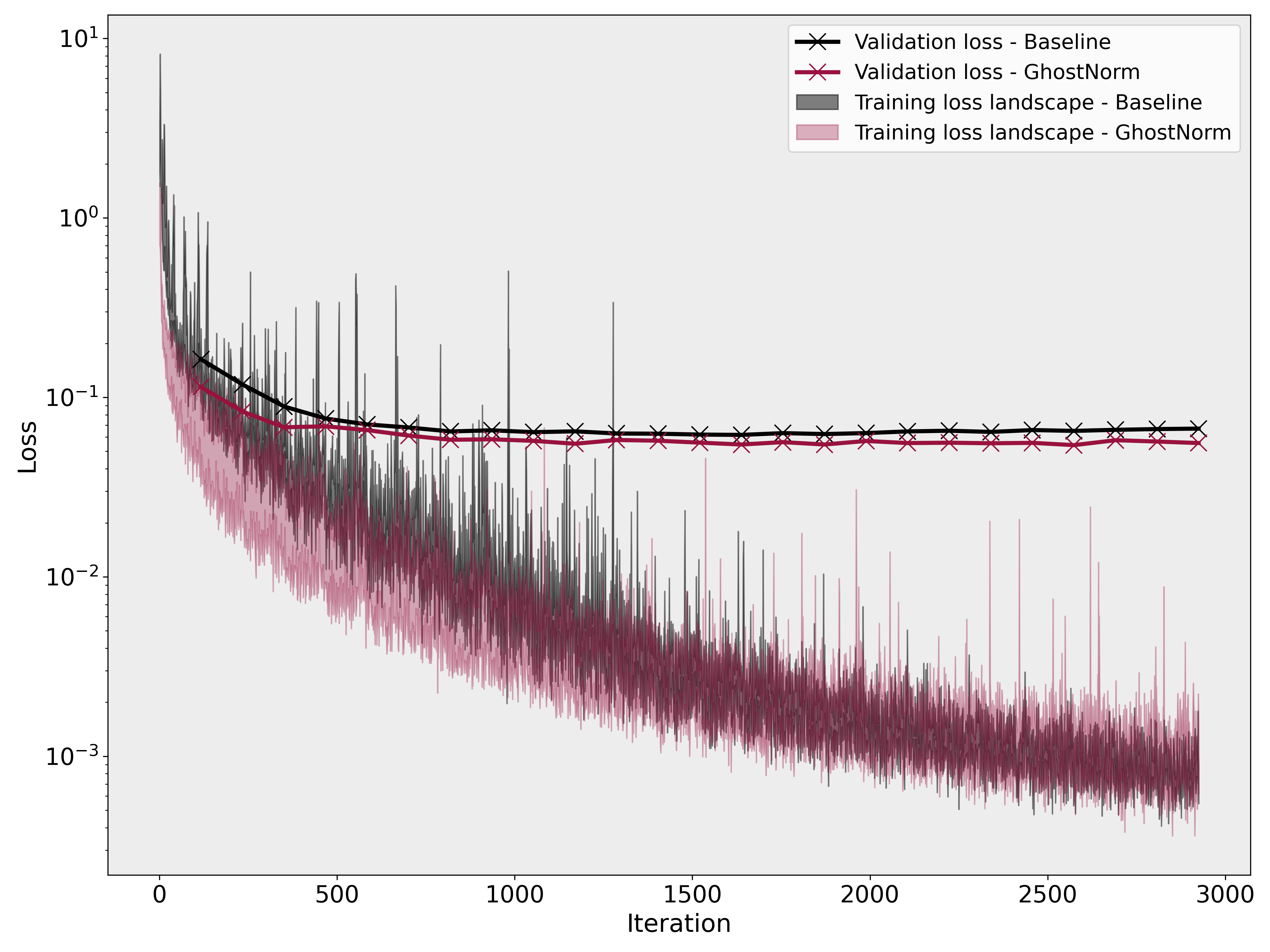}
\end{subfigure}%
\begin{subfigure}{.5\textwidth}
  \centering
  \includegraphics[width=0.8\linewidth]{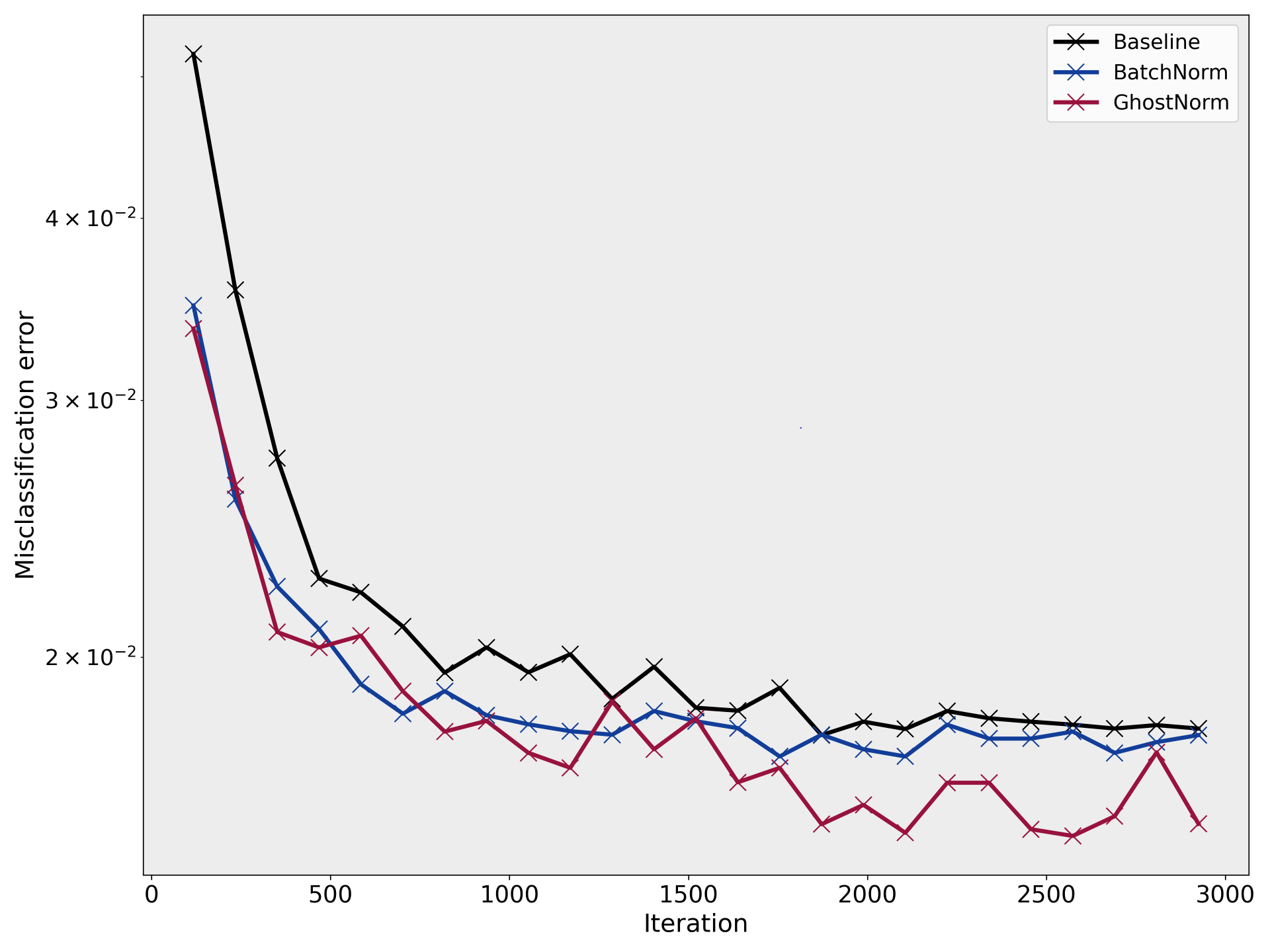}
\end{subfigure}
\caption{Comparison of the loss landscape on MNIST between BatchNorm, GhostNorm, and the baseline.}
% Note that for GhostNorm, a $G_M$ of $8$ is used which can be thought as employing BatchNorm with $8$ GPUs.}
\label{fig:MNIST_loss_landscape}
\end{figure}

\begin{figure}
\centering
\begin{subfigure}{.5\textwidth}
  \centering
  \includegraphics[width=0.8\linewidth]{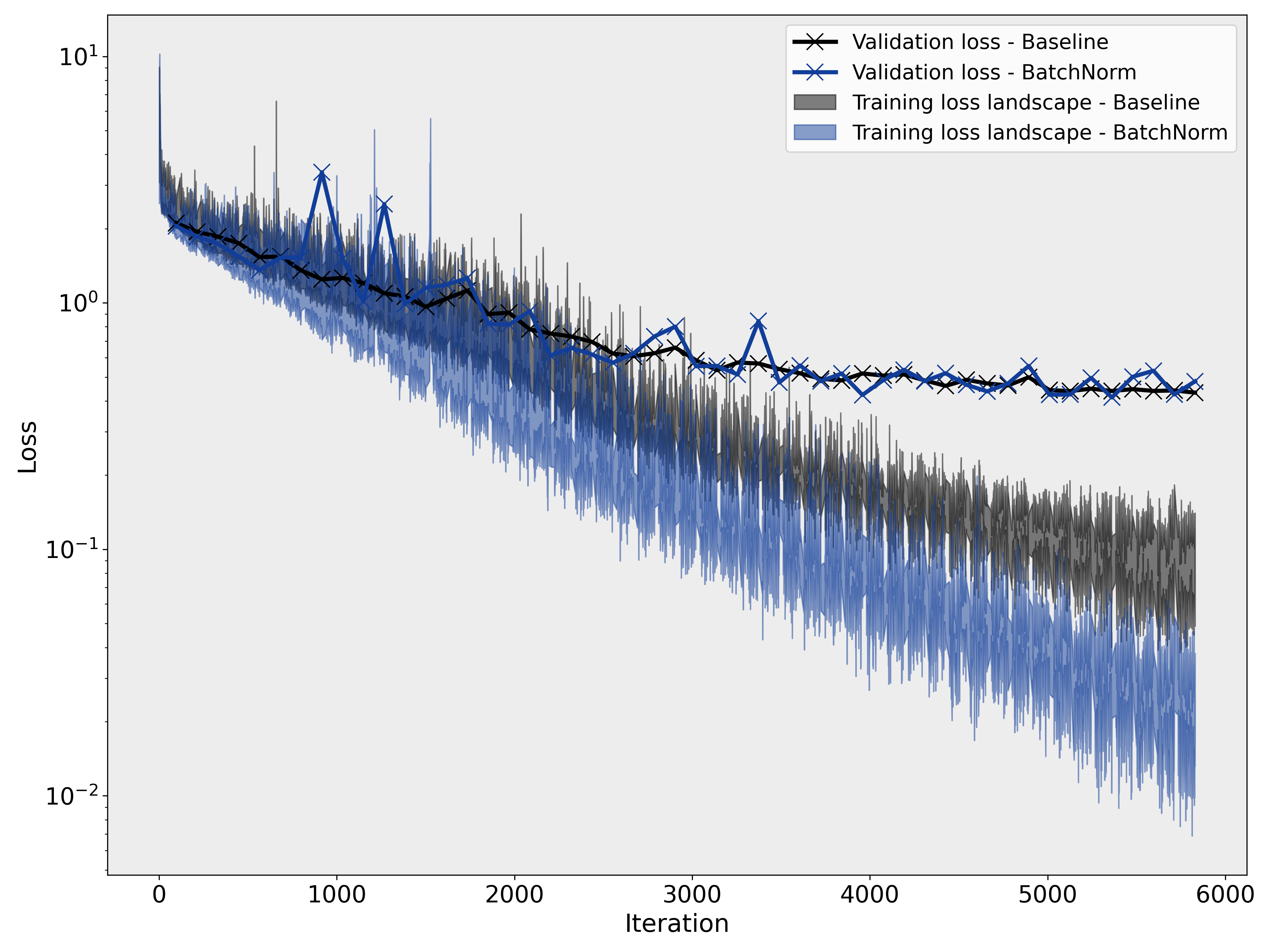}
\end{subfigure}%
\begin{subfigure}{.5\textwidth}
  \centering
  \includegraphics[width=0.8\linewidth]{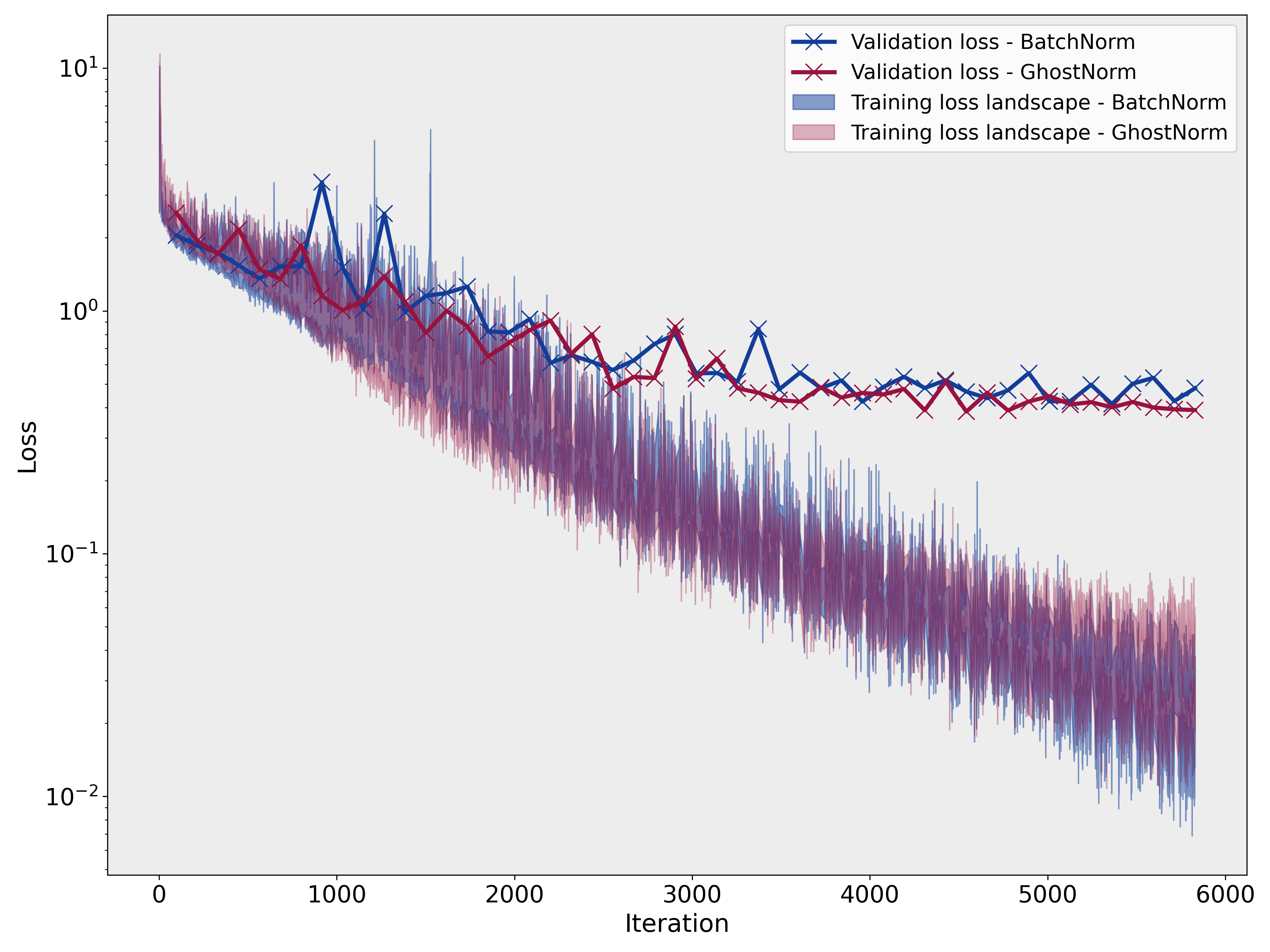}
\end{subfigure}
\vskip\baselineskip
\begin{subfigure}{.5\textwidth}
  \centering
  \includegraphics[width=0.8\linewidth]{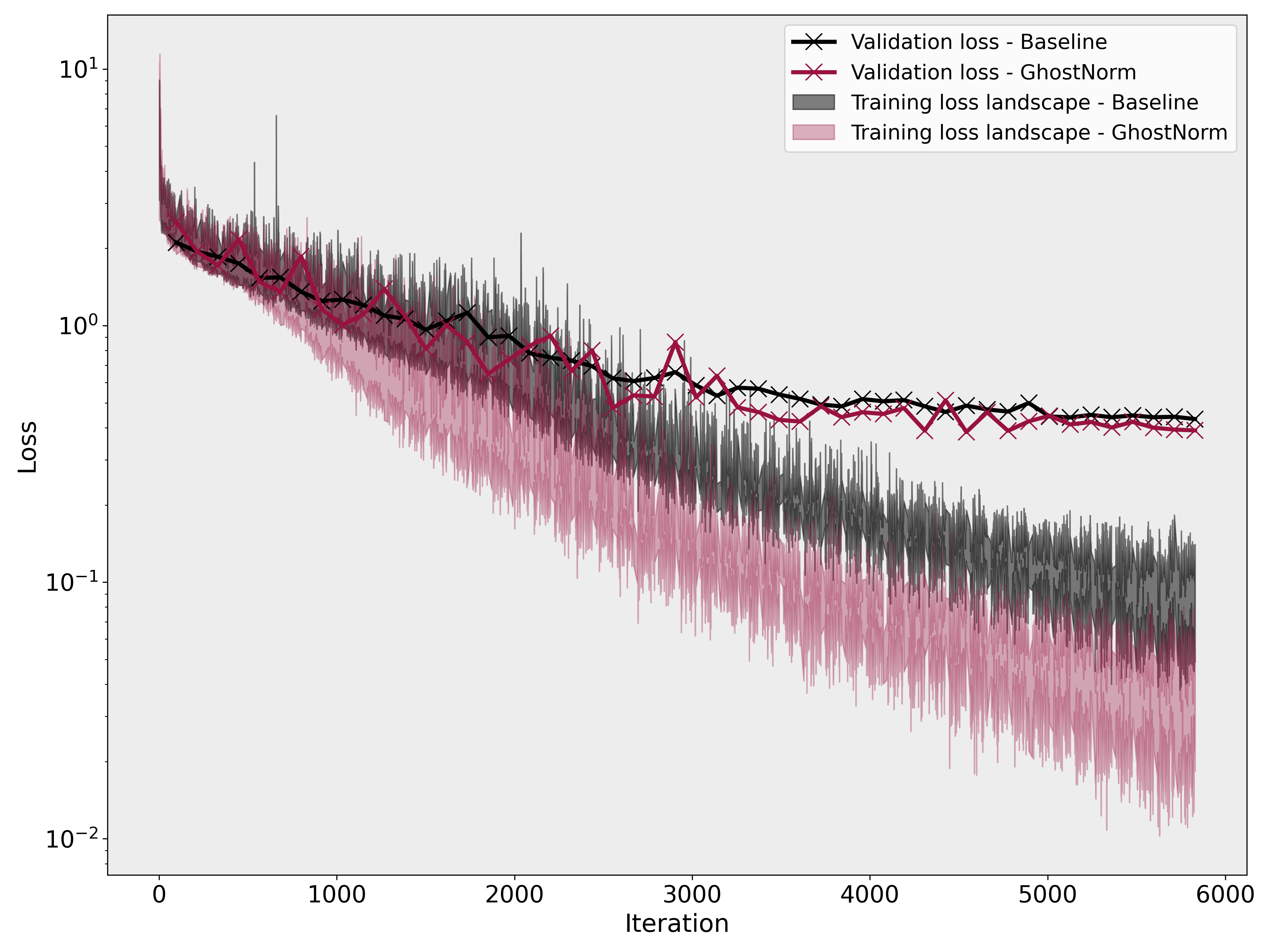}
\end{subfigure}%
\begin{subfigure}{.5\textwidth}
  \centering
  \includegraphics[width=0.8\linewidth]{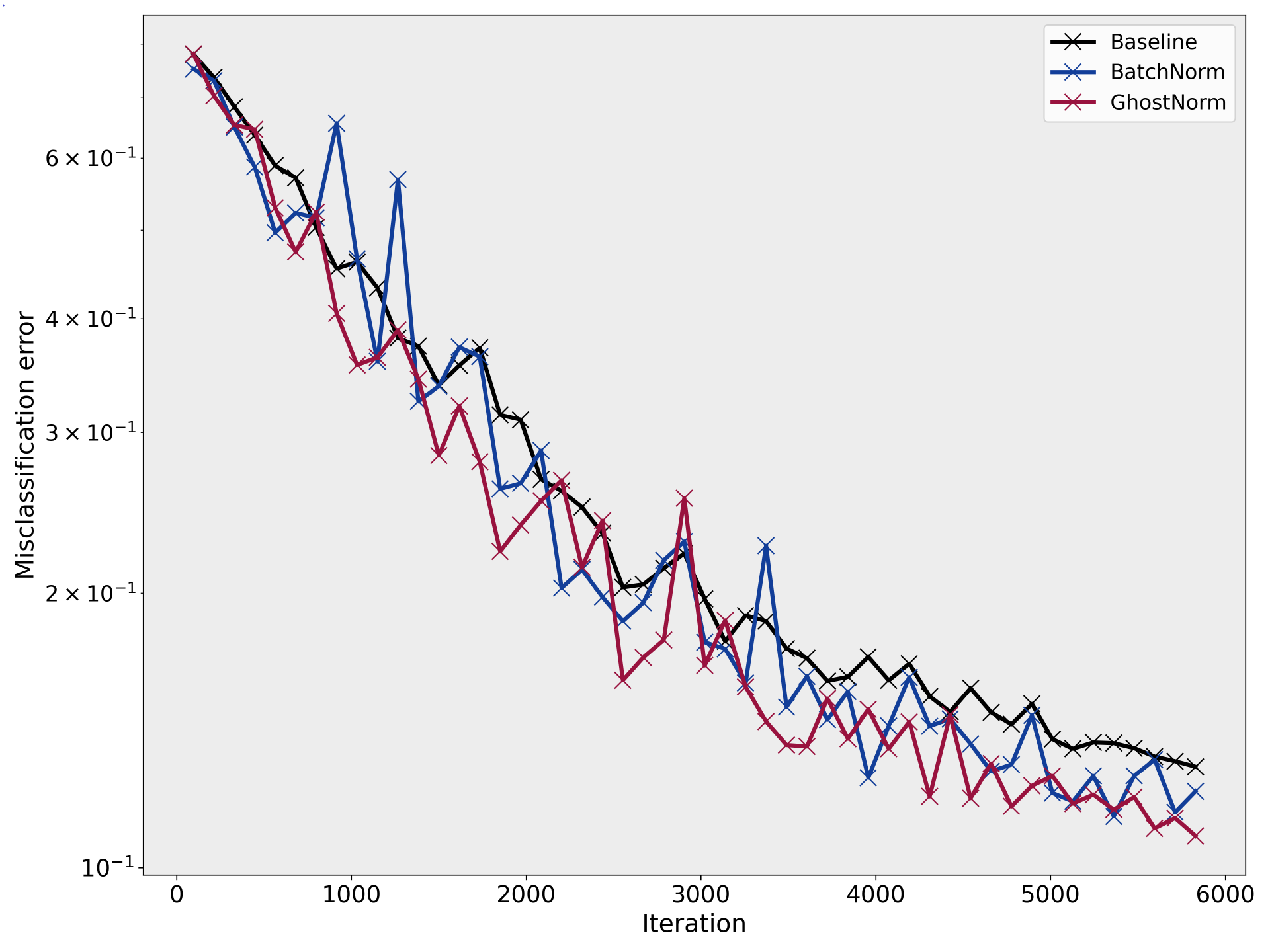}
\end{subfigure}
\caption{Comparison of the loss landscape on CIFAR--10 between BatchNorm, GhostNorm, and the baseline.}
% Note that for GhostNorm, a $G_M$ of $64$ is used.}
% which can be thought as employing GhostNorm with $8$ groups on $8$ GPUs.}
\label{fig:CIFAR10_loss_landscape}
\end{figure}

For both datasets and networks, we observe that the smoothness of the loss landscape deteriorates when GhostNorm is employed. In fact for MNIST, as seen in Figure~\ref{fig:MNIST_loss_landscape}, the loss landscape of GhostNorm bears closer resemblance to our baseline which did not use any normalization technique. For CIFAR--10, this is only observable towards the last epochs of training. In spite of the above observation, we have consistently witness better generalization performances with GhostNorm in almost all of our experiments, even at the extremes wherein $G_M$ is set to $128$, i.e.\ only 4 samples per group. We include further experimental results and discussion in the Appendix. 

% As we increase $G_M$, the smoothness increases, eventually leading to being "too much" To avoid overwhelming the reader, we include further experimental results and discussion in the Appendix. 
% Did not affect 
Our experimental results challenge the often established correlation between a smoother loss landscape and a better generalization performance~\cite{SantTsipIlyaMadr2018,Qiao2019}. Although beyond the scope of our work, a theoretical analysis of the implications of GhostNorm when compared to BatchNorm could potentially uncover further insights into the optimization mechanisms of both normalization techniques.

\subsection{Image classification}
% Our experiments so far have been focused on uncovering more insights around GhostNorm. 
In this section, we further explore the effectiveness of both GhostNorm and SeqNorm on optimization using state-of-the-art (SOTA) methodologies for image classification. Note that the maximum $G_C$ is by architectural design set to $16$, i.e.\ the smallest layer in the network has $16$ channels.

\paragraph{Implementation details}
For both CIFAR--10 and CIFAR--100, we employ a training set of $45,000$ images, a validation set of $5,000$ images (randomly stratified from the training set), and a testing set of $10,000$. 
% For SVHN, only the core set is used which contains approximately $73,000$ training images (of which around $15,000$ are sampled for validation), and $26,000$ testing images. 
The input data were stochastically augmented with horizontal flips, pad-and-crop as well as Cutout~\cite{Devries2017}. We use the same hyperparameter configurations as Cubuk et al.~\cite{Cubuk2019}.
% the choice of optimizer, weight decay, momentum, and initial learning rate
However, in order to speed up optimization, we increase the batch size from 128 to 512, and apply a warmup scheme~\cite{Goyal2017} that increases the initial learning rate by four times in $5$ epochs; thereafter we use the cosine learning schedule. Based on the above experimental settings, we train Wide-ResNet models of $28$ depth and $10$ width~\cite{ZagoKomo2016} for $200$ epochs. Note that since $8$ GPUs are employed, our BatchNorm baselines are equivalent to using GhostNorm with $G_M=8$. Nevertheless, to avoid any confusion, we refer to it as BatchNorm. It's worth mentioning that setting $G_M$ to $8$ on $8$ GPUs is equivalent to using $64$ on $1$ GPU. 

% The best performing model is evaluated on the testing set. For comparative purposes, we also evaluate BatchNorm model. Any model that was evaluated on the testing set was first retrained using both the training and the validation sets.

\paragraph{CIFAR--100} Initially, we turn to CIFAR--100, and tune the hyperparameters $G_C$ and $G_M$ of SeqNorm in a grid-search fashion. The results are shown in Table~\ref{table:cifar100_results}. Both GhostNorm and SeqNorm improve upon the baseline by a large margin ($0.7\%$ and $1.7\%$ respectively). Moreover, SeqNorm surpasses the current SOTA performance on CIFAR--100, which uses a data augmentation strategy, by $0.5\%$~\cite{Cubuk2019}. These results support our hypothesis that sequentially applying GhostNorm and GroupNorm can have an additive effect on improving NN optimization. 

However, the grid--search approach to tuning $G_C$ and $G_M$ is rather time consuming (time complexity: $\Theta(G_C \times G_M)$). Hence, we attempt to identify a less demanding hyperparameter tuning approach. The most obvious, and the one we actually adopt for the next experiment, is to sequentially tune $G_C$ and $G_M$. In particular, we find that first tuning $G_M$, then selecting the largest $g_M \in G_M$ for which the network performs well, and finally tuning $G_C$ with $g_M$ to be an effective approach (time complexity: $\Theta(G_C + G_M)$). Note that by following this approach on CIFAR-100, we still end up with the same best SeqNorm, i.e.\ $G_C=4$ and $G_M=8$.

% we realise that the introduction of two extra hyperparameters (time complexity: $\bigO(G_C \times G_M)$) might not be feasible for most practitioners. Based on the results on CIFAR--100, we reduce the time complexity to $\bigO(G_C + G_M)$ by identifying a hyperparameter tuning strategy for SeqNorm.

\setlength{\tabcolsep}{4pt}
\begin{table}
\begin{center}
\caption{Results on CIFAR--100 by tuning both $G_C$ and $G_M$ in a grid--search fashion. For SeqNorm, we only show the best results for each $G_C$. Both validation and testing performances are averaged over two different runs. Given the same mean performances between two hyperparameter configurations, the one exhibiting less performance variance was adopted.}
\label{table:cifar100_results}
\begin{tabular}{l|cc}
\thickhline
% \noalign{\smallskip}
Normalization & Validation accuracy & Testing accuracy\\
% \noalign{\smallskip}
\hline
BatchNorm           & $80.6\%$ & $82.1\%$ \\
\hline
GhostNorm ($G_M=2$) & $80.9\%$& - \\
GhostNorm ($G_M=4$) & $81.2\%$& - \\
GhostNorm ($G_M=8$) & $\pmb{81.4\%}$& $82.8\%$\\
GhostNorm ($G_M=16$) & $80.3\%$& - \\
\hline
SeqNorm ($G_C=1$, $G_M=4$) & $82.3\%$& - \\
SeqNorm ($G_C=2$, $G_M=4$) & $82.4\%$& - \\
SeqNorm ($G_C=4$, $G_M=8$) & $\pmb{82.5\%}$& $\pmb{83.8\%}$\\
SeqNorm ($G_C=8$, $G_M=8$) & $82.4\%$& - \\
SeqNorm ($G_C=16$, $G_M=8$) & $82.3\%$& - \\
\thickhline
\end{tabular}
\end{center}
\end{table}
\setlength{\tabcolsep}{1.4pt}

\paragraph{CIFAR--10} As the first step, we tune $G_M$ for GhostNorm. We observe that for ${G_M \in (2, 4, 8)}$, the network performs similarly on the validation set at $\approx96.6\%$ accuracy. We choose $G_M=4$ for GhostNorm since it exhibits slightly higher accuracy at $96.7\%$.

Based on the tuning strategy described in the previous section, we adopt $G_M=8$ and tune $G_C$ for values between $1$ and $16$, inclusively. Although the network performs similarly at $\approx96.8\%$ accuracy for ${G_C \in (1, 8, 16)}$, we choose $G_C=16$ as it achieves slightly higher accuracy than the rest. Using the above configuration, SeqNorm is able to match the current SOTA on the testing set~\cite{CubuZophManeVasu+2019}.

% oreover, SeqNorm surpasses the current SOTA performance on CIFAR--100, which uses a data augmentation strategy, by $0.5\%$ [ref].
% I need SVHN results to further elaborate on this as well as the tuning strategy.

\setlength{\tabcolsep}{4pt}
\begin{table}
\begin{center}
\caption{Results on CIFAR--10 based on the sequential tuning of $G_C$ and $G_M$. Both validation and testing performances are averaged over two different runs.}
\label{table:cifar10_results}
\begin{tabular}{l|cc}
\thickhline
% \noalign{\smallskip}
 Normalization & Validation accuracy & Testing accuracy\\
% \noalign{\smallskip}
\hline
BatchNorm           & $96.6\%$ & $97.1\%$ \\
% \hline
% GhostNorm ($G_M=2$) & $96.6\%$& - \\
GhostNorm ($G_M=4$) & $96.7\%$& $97.3\%$ \\
% GhostNorm ($G_M=8$) & $96.6\%$& -\\
% GhostNorm ($G_M=16$) & $96.5\%$& - \\
% \hline
% SeqNorm ($G_C=1$, $G_M=8$) & $96.8\%$& - \\
% SeqNorm ($G_C=2$, $G_M=8$) & $96.7\%$& - \\
% SeqNorm ($G_C=4$, $G_M=8$) & $96.6\%$& - \\
% SeqNorm ($G_C=8$, $G_M=8$) & $96.8\%$& - \\
SeqNorm ($G_C=16$, $G_M=8$) & $96.8\%$& $\pmb{97.4\%}$ \\
\thickhline
\end{tabular}
\end{center}
\end{table}
\setlength{\tabcolsep}{1.4pt}

\subsection{Negative Results}
A number of other approaches were adopted in conjunction with GhostNorm and SeqNorm. These preliminary experiments did not surpass the BatchNorm baseline performances on the validation sets (most often than not by a large margin), and are therefore not included in detail. Note that given a more elaborate hyperparameter tuning phase, these approaches may had otherwise succeeded. 

In particular, we have also experimented with placing GhostNorm and GroupNorm in reverse order for SeqNorm (in retrospect, this could have been expected given what we describe in Section~\ref{sec:effects_gn}), and have also experimented with augmenting SeqNorm and GhostNorm with weight standardisation~\cite{Qiao2019} as well as by computing the variance of batch statistics on the whole input tensor [ref]. Finally, for all experiments, we have attempted to tune networks with only GroupNorm~\cite{WuHe2018} but the networks were either unable to converge or they achieved worse performances than the BatchNorm baselines.

\section{Conclusion}
It is generally believed that the cause of performance deterioration of BatchNorm with smaller batch sizes stems from it having to estimate layer statistics using smaller sample sizes~\cite{Sergey2017,WuHe2018,YanWanZhanZhan+2020}. In this work we challenged this belief by demonstrating the effectiveness of GhostNorm on a number of different networks, learning policies, and datasets. For instance, when using super--convergence on CIFAR--10, GhostNorm performs better than BatchNorm, even though the former normalizes the input activations using $4$ samples whereas the latter uses all $512$ samples. By providing novel insight on the source of regularization in GhostNorm, and by introducing a number of possible implementations, we hope to inspire further research into GhostNorm. 

Moreover, based on the understanding developed while investigating GhostNorm, we introduce SeqNorm and follow up with empirical analysis. Surprisingly, SeqNorm not only surpasses the performances of BatchNorm and GhostNorm, but even challenges the current SOTA methodologies on both CIFAR--10 and CIFAR--100 that employ data augmentation strategies~\cite{Cubuk2019,CubuZophManeVasu+2019}. Finally, we also describe a tuning strategy for SeqNorm that provides a faster alternative to the traditional grid--search approach.

\bibliographystyle{unsrt}  
\bibliography{references}  %%% Remove comment to use the external .bib file (using bibtex).

\end{document}